\documentclass[sigconf]{acmart}

\AtBeginDocument{%
  \providecommand\BibTeX{{%
    \normalfont B\kern-0.5em{\scshape i\kern-0.25em b}\kern-0.8em\TeX}}}

\copyrightyear{2023}
\acmYear{2023}
\setcopyright{acmlicensed}\acmConference[MM '23]{Proceedings of the 31st ACM International Conference on Multimedia}{October 29-November 3, 2023}{Ottawa, ON, Canada}
\acmBooktitle{Proceedings of the 31st ACM International Conference on Multimedia (MM '23), October 29-November 3, 2023, Ottawa, ON, Canada}
\acmPrice{15.00}
\acmDOI{10.1145/3581783.3611951}
\acmISBN{979-8-4007-0108-5/23/10}

\usepackage{algorithm}
\usepackage{algorithmic}

\usepackage{subfig}
\usepackage{mathrsfs}

\usepackage{amssymb}
\usepackage{amsmath}
\usepackage{dsfont}
\usepackage{booktabs}
\usepackage{epstopdf}
\usepackage{multirow}
\usepackage{comment}
\usepackage{endnotes}
\usepackage{hyperref}
\usepackage{bibentry}
\usepackage{amsmath}
\usepackage{dsfont}
\usepackage{amsfonts}

\usepackage{amssymb}
\usepackage{bbm}
\usepackage{bbding}
\usepackage{makecell}
\usepackage[normalem]{ulem}
\usepackage{balance}
\useunder{\uline}{\ul}{}

\begin{document}
\title{DealMVC: Dual Contrastive Calibration for Multi-view Clustering}
\author{Xihong Yang}
\email{xihong_edu@163.com}
\affiliation{%
  \institution{National University of Defense Technology}
  \city{Changsha}
  \state{Hunan}
  \country{China}
}

\author{Jin Jiaqi}
\affiliation{%
  \institution{National University of Defense Technology}
  \city{Changsha}
  \state{Hunan}
  \country{China}
}

\author{Siwei Wang}
\affiliation{%
  \institution{Intelligent Game and Decision Lab}
  \city{Beijing}
  \country{China}
}

\author{Ke Liang}
\author{Yue Liu}
\affiliation{%
  \institution{National University of Defense Technology}
  \city{Changsha}
  \state{Hunan}
  \country{China}
}

\author{Yi Wen}
\author{Suyuan Liu}
\author{Sihang Zhou}
\affiliation{%
  \institution{National University of Defense Technology}
  \city{Changsha}
  \state{Hunan}
  \country{China}
}

\author{Xinwang Liu}
\authornote{Corresponding author}
\author{En Zhu}
\authornotemark[1]
\affiliation{%
  \institution{National University of Defense Technology}
  \city{Changsha}
  \state{Hunan}
  \country{China}
}

\renewcommand{\shortauthors}{Xihong Yang et al.}

\begin{abstract}

Benefiting from the strong view-consistent information mining capacity, multi-view contrastive clustering has attracted plenty of attention in recent years. However, we observe the following drawback, which limits the clustering performance from further improvement. The existing multi-view models mainly focus on the consistency of the same samples in different views while ignoring the circumstance of similar but different samples in cross-view scenarios. To solve this problem, we propose a novel \textbf{D}ual contrastiv\textbf{e} c\textbf{al}ibration network for \textbf{M}ulti-\textbf{V}iew \textbf{C}lustering (DealMVC). Specifically, we first design a fusion mechanism to obtain a global cross-view feature. Then, a global contrastive calibration loss is proposed by aligning the view feature similarity graph and the high-confidence pseudo-label graph. Moreover, to utilize the diversity of multi-view information, we propose a local contrastive calibration loss to constrain the consistency of pair-wise view features. The feature structure is regularized by reliable class information, thus guaranteeing similar samples have similar features in different views. During the training procedure, the interacted cross-view feature is jointly optimized at both local and global levels. In comparison with other state-of-the-art approaches, the comprehensive experimental results obtained from eight benchmark datasets provide substantial validation of the effectiveness and superiority of our algorithm. We release the code of DealMVC at https://github.com/xihongyang1999/DealMVC on GitHub.

\end{abstract}

\ccsdesc[500]{Theory of computation~Unsupervised learning and clustering}
\ccsdesc[500]{Computing methodologies~Cluster analysis}

\keywords{Multi-view Clustering, Contrastive Learning}

\maketitle

\section{INTRODUCTION}

Multi-view clustering (MVC) has attracted increasing attention in recent years. MVC is a fundamental task to reveal the semantic information and divides the data into several disjoint groups. The existing multi-view clustering algorithms could be roughly divided into two categories, including conventional and deep algorithms. 

The conventional MVC algorithms optimize the formulation of traditional machine learning methods. These MVC methods can be basically categorized into four classes: non-negative matrix factorization (NMF) \cite{mvc_mf,wenjie_mf,mvc_mf2,mvc_mf3}, multiple kernel clustering (MKC) \cite{one,simplemkkm,incomplete,efficient}, graph-based clustering\cite{suyuan_AAAI,suyuan_TNNLS,siwei_nips} and the subspace clustering methods \cite{ZHOU_1,ZHOU_2,zhang2020adaptive,zhu2019multi}. Specifically, NMF utilizes matrix factorization for the multi-view data. MKC algorithms apply the kernel matrix from the predefined kernel matrices through the multiple kernel learning framework. Moreover, graph-based clustering exploits the multi-view data with a unified graph structure. Besides, the subspace clustering methods focus on learning consistent subspace representations. However, those traditional multi-view clustering methods suffer from poor representation extraction capacity and high computation complexity, thus limiting the clustering performance.

\begin{figure}[t]
\centering
\includegraphics[scale=0.32]{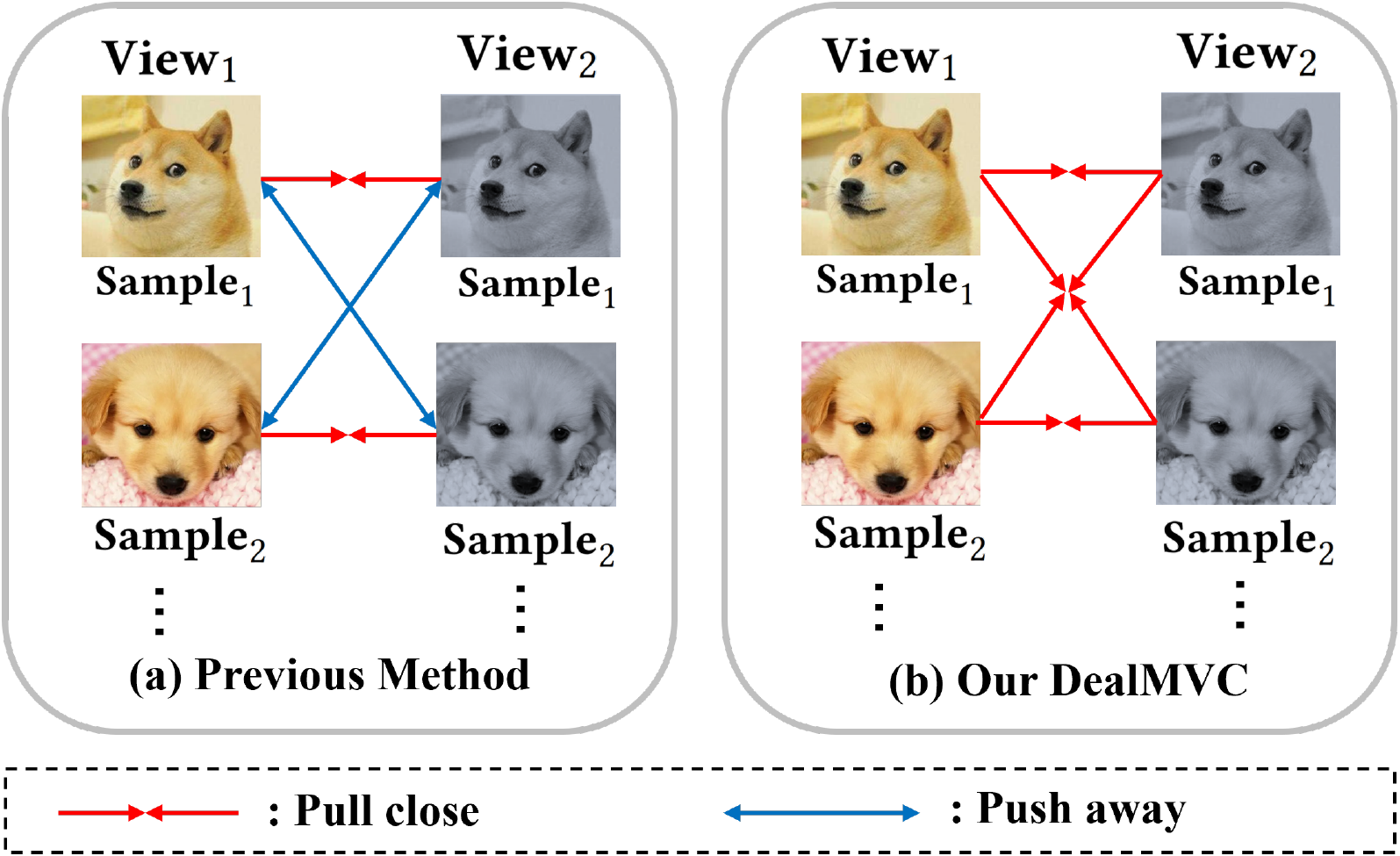}
\caption{An example illustration of the motivation. Most multi-view clustering methods focus on keeping the consistent of the same sample in different views, i.e., pulling $\textbf{Sample}_1$ in $\textbf{View}_1$ and $\textbf{Sample}_1$ in $\textbf{View}_2$ close, while neglecting the consistent of similar but different samples, i.e., $\textbf{Sample}_1$ in $\textbf{View}_1$ and $\textbf{Sample}_2$ in $\textbf{View}_2$.}
\label{motivation}  
\end{figure}

Thanks to the outstanding representation extraction ability, deep clustering methods have been proposed to alleviate the above issue \cite{dmvc1,dmvc3,dmvc4,Completer,EAMC,SiMVC}. Deep graph-based algorithms \cite{graph_m1,graph_m2} utilized affinity matrices to directly cluster the multi-view data. Moreover, the adversarial multi-view algorithms \cite{adver_m1,adver_m2} use generators and discriminators to align feature distribution for multi-view data. More recently, contrastive learning has become an effective fashion for many fields. COMPLETER \cite{Completer} learned the informative and consistent representation of multi-view data through contrastive learning. MFLVC \cite{MFLVC} learns different levels of features with contrastive strategy.

\begin{figure*}
\centering
\includegraphics[scale=0.5]{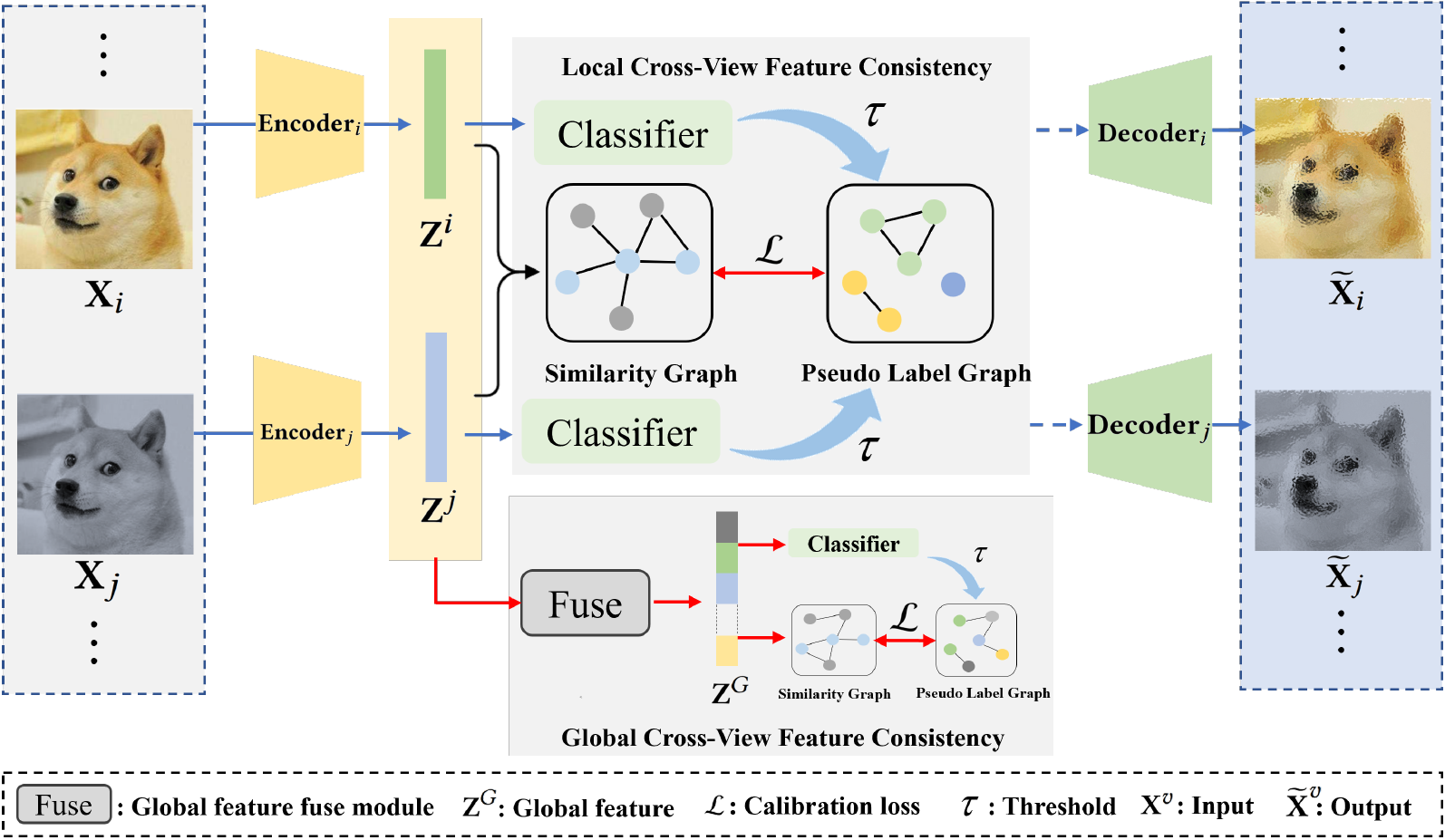}
\caption{Illustration of the dual contrastive calibration network. In our method, we first obtain the view feature $\textbf{Z}$ by the encoder network. After that, an adaptive fusion mechanism is designed to obtain the global view feature. Detailed descriptions are shown in Fig. \ref{global_fusion}. Then, we align the view feature similarity graph and the high-confidence pseudo-label graph to keep the consistency of the different but similar samples in the cross-view scenario. The feature view structure is regularized by the reliable pseudo labels, which are interacted and jointly optimized by the network, thus improving the multi-view clustering performance.}
\label{overall}   
\end{figure*}

Despite achieving promising performance, the majority of existing algorithms tend to emphasize the consistency of identical samples across different views, inadvertently overlooking the potential of similar yet distinct samples in cross-view scenarios. As demonstrated in Figure \ref{motivation} (a), prevalent contrastive multi-view clustering methods primarily concentrate on maintaining consistency among identical samples, such as $sample_1$ in $View_1$ and $sample_1$ in $View_2$. However, a prevalent situation involves the likeness between samples from diverse views, for instance, $sample_1$ in $View_1$ and $sample_2$ in $View_2$. Addressing the challenge of preserving the consistency for akin samples within cross-view in unsupervised settings constitutes a complex problem.

\begin{figure}
\centering
\scalebox{0.32}{
\includegraphics{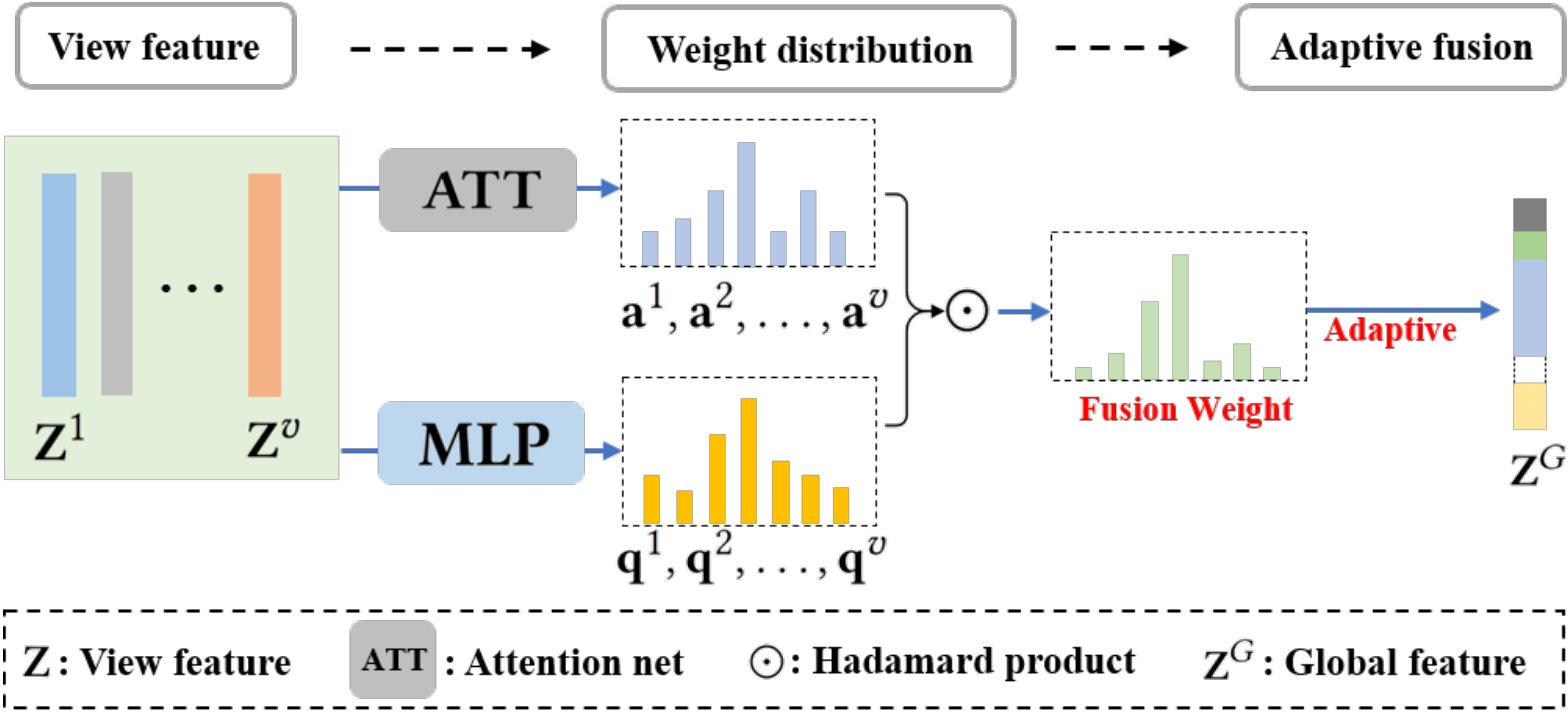}}
\caption{Illustration of the Adaptive Global Fusion Mechanism. In our proposed algorithm, we first obtain two distributions $\textbf{a} \in \mathbb{R}^V$ and $\textbf{q} \in \mathbb{R}^V$ through attention network and Multilayer Perceptron (MLP) network, respectively. After that, we combine the attention weight $\textbf{a}$ and the learnable view sampling probability $\textbf{q}$ by the regulatory factor $\textbf{r} \in \mathbb{R}^V$. When the attention vector and the view sampling probability have a similar distribution, $\textbf{r}$ has a high value to effectively extract the important view information. Moreover, the global view feature $\textbf{Z}^G$ could be obtained by the adaptive weight vector $\textbf{w}$.}
\label{global_fusion}
\end{figure}

To deal with the above issues, in this paper, we propose a novel dual contrastive calibration network for multi-view clustering, termed DealMVC. The framework of our DealMVC is shown in Fig. \ref{overall}. We first design an adaptive global fusion mechanism to obtain the global cross-view feature in two manners, i.e., an attention mechanism and a learnable view sampling network. The detailed description is shown in Fig.\ref{global_fusion}. After that, we design a global contrastive calibration loss to guarantee the samples with similar features, which aligns the view feature similarity graph and the high-confidence pseudo-label graph. To further mine the diversity of the multi-view information, we design a local contrastive calibration loss for pair-wise views. The structure of the cross-view feature is consistent with the reliable class information. During the training procedure, the view feature and the high-confidence pseudo labels are jointly optimized and interacted with the dual contrastive calibration loss. Compared with the existing multi-view clustering methods, extensive experiments on eight datasets have demonstrated the effectiveness and the superiority of our proposed DealMVC.

The main contributions of this paper are summarized as follows.

\begin{itemize}
\item We propose a novel deep contrastive multi-view clustering algorithm termed DealMVC. The dual contrastive calibration mechanism keeps the consistency of similar but different samples in the cross-view scenario.

\item By aligning the view feature similarity graph and the high-confidence pseudo-label graph, the view feature structure is constrained by reliable class information.

\item Extensive experimentation across eight benchmark datasets serves to highlight both the superiority and efficiency of the proposed DealMVC method. Furthermore, the effectiveness of our approach is substantiated by ablation studies and visualization experiments.

\end{itemize}

\section{RELATED WORK}

\subsection{Multi-view Clustering }

In recent times, Multi-view Clustering (MVC) \cite{wenyi, wan1, wan2, wan3, wan4} has garnered significant attention. Existing MVC algorithms can be broadly categorized into two groups: traditional multi-view clustering methods and deep multi-view clustering methods. The traditional approaches encompass four main categories. (1) Matrix factorization-based algorithms \cite{mvc_mf,mvc_mf2}. Non-negative matrix factorization techniques aim to uncover a shared latent factor to extract information from multi-view data. \cite{mvc_mf3} introduced a shared clustering indicator matrix within the multi-view context. (2) Kernel learning-based MVC employs predefined kernels to handle diverse views. Efforts are made to devise a unified kernel by linear or non-linear combinations of predefined kernels. (3) Graph-based MVC \cite{CMC,CGD,graph_mvc} utilize multi-view data to construct graphs that preserve sample structures. (4) Subspace-based MVC focuses on consistent subspace representation learning across multiple views. In \cite{subspace}, a diversity-induced mechanism was designed for multi-view subspace clustering. However, these traditional methods tend to capture shallow representations of multi-view data, thus limiting the discriminative capacity of the acquired representations.

Benefiting from robust feature representations, deep networks have the capacity to extract more refined feature representations. In recent years, deep multi-view clustering methods have been introduced \cite{dmvc1,dmvc4,dmvc3,jiaqi_cvpr}. These methods can generally be categorized into two groups: one-stage algorithms \cite{DAMC,EAMC} and two-stage algorithms \cite{MultiVAE, Completer}. Through deep multi-view clustering, latent cluster patterns within multi-view data can be effectively uncovered.

\begin{table}[]
\centering
\caption{Basic notations used in the whole paper.}
\begin{tabular}{@{}cc@{}}
\toprule
\textbf{Notation} & \textbf{Meaning}                  \\ \midrule
$\textbf{X}^v$                 & The data matrix of $v$-th view       \\
$\textbf{Z}^v$                  & The feature of $v$-th view        \\
$\textbf{Z}^G$                  & The global feature of all views     \\
$V$               &  the view number of the multi-view data set  \\ 
$\textbf{a}$        &  the attention vector  \\
$\textbf{S}$                  & The view feature similarity graph \\
$\textbf{W}$                 & The high-confidence pseudo-label graph          \\
$\textbf{q}$          & learnable view sampling probability vector  \\
$\textbf{p}$                  & The pseudo labels               \\
$\textbf{r}$                  & The regulatory factor             \\ \bottomrule
\end{tabular}
\label{NOTATION_TABLE}
\end{table}

\subsection{Contrastive Learning}
The research domains of computer vision \cite{wan_1, wan_2, he1, he2,xihong,tan1, tan2, tan3, tan4, sun1, sun2, sun3,xiaochang} and graph learning \cite{DCRN,IDCRN,CCGC, HSAN, GCC-LDA, SCGC, MGCN, Dink_net} hold significant importance. Contrastive learning, renowned for its robust intrinsic supervision information extraction capabilities, has garnered substantial interest. The central principle of contrastive learning revolves around maximizing the similarity between positive samples while minimizing the similarity among negative samples in the latent space. Noise contrastive estimation (NCE) \cite{nce1,nce2} was introduced, followed by InfoNCE \cite{infonce}, aimed at discerning different views of a sample amid others. Moco \cite{MOCO} and SimCLR \cite{SIMCLR} subsequently learned image-wise features by pulling together positive sample pairs and pushing apart negative sample pairs. In the context of multi-view clustering, several contrastive learning methods have emerged \cite{CMC,MVGRL,MFLVC}. CMC \cite{CMC} devised a contrastive multi-view coding framework to extract underlying semantic information. MVGRL \cite{MVGRL} employed the graph diffusion matrix to generate augmented graphs, subsequently establishing a multi-view contrastive mechanism for downstream tasks. More recently, MFLVC \cite{MFLVC} introduced two objectives for multi-view clustering on high-level features and pseudo labels, employing contrastive learning.

\section{METHODOLOGY}

This section proposes a novel dual contrastive calibration network for multi-view clustering, termed DealMVC. The framework is shown in Fig. \ref{overall}. Moreover, detailed notations and corresponding meanings are summarized in Table \ref{NOTATION_TABLE}.

\subsection{AutoEncoder Module}

Recently, many works \cite{autoencoder1,autoencoder2} demonstrate that autoencoder is widely used in unsupervised scenarios. Motivated by the effectiveness, we design the autoencoder module to project the features to a customizable feature space. In this way, MVC could utilize the semantics across all views to improve the clustering performance. To be specific, let $\textbf{Z}^v = f(\textbf{X}^v;\theta^v)$ and $ \widetilde{\textbf{X}}^v = g(\textbf{Z}^v; \phi^v)$ denote the encoder and decoder process, respectively. $\theta^v$ and $\phi^v$ are the network parameters. Therefore, we design the reconstruction loss of all $v$-view between the input $\textbf{X}^v$ and the output $\widetilde{\textbf{X}}^v$ as:

\begin{equation} 
\mathcal{L}_R = \sum_{v=1}^V  \left \| \textbf{X}^v - \widetilde{\textbf{X}}^v \right \|_F^2.
\label{auto_loss}
\end{equation}

\subsection{Adaptive Global Fusion Module}

In this subsection, we propose an adaptive global fusion mechanism to obtain the global cross-view feature. Based on $\textbf{Z}^v = f(\textbf{X}^v;\theta^v)$, we design a regulatory factor to adaptive fuse $v$ views. Let $\textbf{w}= (w_1, w_2, \dots, w_v)$ as the weight vector for each view. The weight vector $\textbf{w} \in \mathbb{R}^V$ is obtained in two manners, i.e., multi-head attention and learnable view sample probability vector.

Concretely, inspired by the success of the attention mechanism \cite{attention_1}, we design an attentive network to capture the important view information under the multi-view scenario. The attention vector $\textbf{a} \in \mathbb{R}^V$ for each view could be calculated as:
\begin{equation} 
\begin{aligned}
\hat{\textbf{Z}} &= ATT(norm(\textbf{Z})),\\
 \textbf{a} &= FFN(\hat{\textbf{Z}}),
\end{aligned}
\label{attention_opt}
\end{equation}
where $norm$ is the normalization before the attention network (ATT), and FFN denotes the feed-forward network. The attention mechanism could mine the important view information. In this way, the fused global view could preserve important semantic patterns.

Meanwhile, to avoid the training bias caused by the attention mechanism, we utilize a learnable view sampling probability vector $\textbf{q} \in \mathbb{R}^V$ to dynamically fuse view during the training process. To be specific, we firstly initialize the $\textbf{q}$ with the uniform distribution. Then, we optimize $\textbf{q}$ as follows:
\begin{equation} 
\begin{aligned}
\textbf{q} = h(\textbf{q},\gamma),
\end{aligned}
\label{mlp_opt}
\end{equation}
where $h(\cdot;\gamma)$ is the view probability network. Here, we adopt Multilayer Perceptron (MLP) network as the backbone. The sampling probability is dynamically adjusted during the training procedure according to the network. 


After that, we propose a regulatory factor to combine the attention mechanism and the learnable view sampling probability. The regulatory factor $\textbf{r} \in \mathbb{R}^V$ could be defined as:
\begin{equation} 
\begin{aligned}
\textbf{r} = \textbf{a} \odot \textbf{q},
\end{aligned}
\label{regulatory_factor}
\end{equation}
where $\odot$ represents the Hadamard product. Through this approach, the merged probability for each view is directly tied to the attention vector and the view sampling probability. In other words, the regulatory factor for adjusting view fusion is more pronounced when there is a greater similarity between the distributions of the attention vector and the view sampling probability.

Based on the regulatory factor, we design an adaptive fusion strategy to obtain the global view feature. Specifically, the weight vector $\textbf{w}$ is adjusted by the regulatory factor $\textbf{r}$ with Hadamard product. The fused global view feature $\textbf{Z}^G$ could be formulated as:
\begin{equation} 
\begin{aligned}
\textbf{w}^{t+1} &= \textbf{r} \odot \textbf{w}^t,\\
\textbf{Z}^G &= \sum_{v=1}^V \textbf{Z}^v \textbf{w}^v,
\end{aligned}
\label{fuse_weight}
\end{equation}
where $t$ is the number of iteration. Different from the previous view fusion strategies, our view fusion strategy is more comprehensive and reliable. The reasons we analyze are as follows. 1) The adaptive weight vector $\textbf{w}$ is contains two aspects information. The learnable view sampling network and the attention mechanism could capture the important view information in two manners, which can be jointly optimized by the network. 2) With the adjusting of the regulatory factor $\textbf{r}$, the main view information is concerned when the probability distribution is similar in two fusion strategies.

\subsection{Dual Contrastive Calibration Module}

In this subsection,  we design a dual contrastive calibration module to keep the consistency of similar but different samples in the cross-view scenario on both global and local view levels. The features $\{\textbf{Z}^v\}_{v=1}^V$ of $v$ views could be obtained by the encoder network $f(\cdot;\theta^v)$.

Based on the fusion cross-view feature $\textbf{Z}^G \in \mathbb{R}^{N \times D}$, we firstly obtain the pseudo labels $\textbf{p}^G \in \mathbb{R}^{N \times K}$ of the global cross-view feature:
\begin{equation} 
\begin{aligned}
\textbf{p}^G = Classify(\textbf{Z}^G),  
\end{aligned}
\label{global_pseudo_label}
\end{equation}
where $Classify(\cdot)$ is the classification head. Then, we construct the global pseudo-label graph $\textbf{W}^{G} \in \mathbb{R}^{N \times N}$ by:
\begin{equation} 
\begin{aligned}
\textbf{W}_{ij}^{G}&=\left\{\begin{array}{ll}
1 & \text { if } i=j, \\
p^{G}_{i} \cdot p^{G}_{j} & \text { if } i \neq j \text { and } p^{G}_{i} \cdot p^{G}_{j} \geq \tau, \\
0 & \text { otherwise },
\end{array}\right.
\end{aligned}
\label{glo_pse}
\end{equation}
where $\tau$ is the threshold. The diagonal elements of $\textbf{W}^{G}$ represent the class probabilities of the same samples in the cross-view feature. For other elements, if the calculated similarity is lower than the threshold $\tau$, the samples are not connected in the pseudo-label graph. After that, we construct global cross-view feature graph $\textbf{S}^{G} \in \mathbb{R}^{N \times N}$:

\begin{equation} 
\begin{aligned}
\textbf{S}^{G} =\frac{\left \langle \textbf{Z}^G, (\textbf{Z}^G)^T  \right \rangle}{\left \| \textbf{Z}^G \right \|_2 \left \| \textbf{Z}^G  \right \|_2},  
\end{aligned}
\label{similar}
\end{equation}
where $\left \langle \cdot , \cdot \right \rangle $ denotes the inner product. To keep the consistency of similar but different samples for the global cross-view feature, we align the global cross-view feature similarity graph and the high-confidence pseudo-label graph as follows:
\begin{equation} 
\begin{aligned} 
\mathcal{L}_{global} = {-\textbf{W}_{ii}^{G}log\left ( \frac{exp{(\textbf{S}_{ii}^{G})}}{\sum exp({\textbf{S}_{ij}^{G}})} \right ) } \\
- \sum_{j=1,j \neq i}^N \textbf{W}_{ij}^{G} log \left ( \frac{exp{(\textbf{S}_{ij}^{G})}}{\sum exp({\textbf{S}_{ij}^{G}})}\right ). 
\end{aligned}
\label{global_con}
\end{equation}

The first term in Eq.\eqref{global_con} pushes the same samples to close in the cross-view feature. The second term forces similar but different samples to have the same clusters. During the training procedure, the pseudo-label graph serves as guidance to train the cross-view feature graph.

\begin{algorithm}[t]
\small
\caption{Dual Contrastive Calibration Network (DealMVC)}
\label{ALGORITHM}
\flushleft{\textbf{Input}: The multi-view raw features $\textbf{X}$}; the interation number $I$ \\
\flushleft{\textbf{Output}: The clustering result \textbf{R}.} 
\begin{algorithmic}[1]
\FOR{$i=1$ to $I$}
\STATE Obtain the view feature $\textbf{Z}$ by the encoder network.
\STATE Fuse the feature to obtain the global cross-view feature by the adaptive fusion module.
\STATE Construct the high-confidence pseudo-label graph and the cross-view feature similarity graph for local and global cross-view scenario.
\STATE Aligning the constructed graph with the contrastive calibration loss.
\STATE Calculate the loss by Eq.\eqref{total_Loss}
\STATE Update model by minimizing $\mathcal{L}$ with Adam optimizer.
\ENDFOR
\STATE \textbf{return} \textbf{R}
\end{algorithmic}
\end{algorithm}

To further utilize the diversity of multi-view information, we design a local contrastive calibration loss for pair-wise view features. Given the view features $\{\textbf{Z}^v\}_{v=1}^V$, we obtain the pseudo labels $\{\textbf{p}^v\}_{v=1}^V \in \mathbb{R}^{N \times K}$ for each view by:
\begin{equation} 
\begin{aligned}
\textbf{p}^v = Classify(\textbf{Z}^v),  
\end{aligned}
\label{pseudo_label}
\end{equation}
where $Classify(\cdot)$ is the classification head. Then, we construct the pseudo-label graph by calculating the similarity matrix $\textbf{W}^{mn} \in \mathbb{R}^{N \times N}$ of any two view pseudo labels $\textbf{p}^m$ and $\textbf{p}^n$, which can be presented as:
\begin{equation} 
\begin{aligned}
\textbf{W}_{ij}^{mn}=\left\{\begin{array}{ll}
1 & \text { if } i=j, \\
p^{mn}_{i} \cdot p^{mn}_{j} & \text { if } i \neq j \text { and } p^{mn}_{i} \cdot p^{mn}_{j} \geq \tau, \\
0 & \text { otherwise },
\end{array}\right.  
\end{aligned}
\label{pse_sim}
\end{equation}
where $m, n \in [1, V]$. Similar to $\textbf{W}^G$, the diagonal elements of $\textbf{W}^{mn}$ represent the class probabilities of the same samples in different views. For other elements, the connection in the pseudo-label graph is constructed when the similarity is higher than the threshold $\tau$.

To further construct the view feature graph, we calculate the similarity matrix $\textbf{S}^{mn} \in \mathbb{R}^{N \times N}$ between any two view features $\textbf{Z}^m$ and $\textbf{Z}^n$:
\begin{equation} 
\begin{aligned}
\textbf{S}_{ij}^{mn} = \frac{\left \langle \textbf{Z}_i^m, \textbf{Z}_j^n  \right \rangle }{\left \| \textbf{Z}_i^m \right \|_2 \left \| \textbf{Z}_j^n  \right \|_2},
\end{aligned}
\label{similar}
\end{equation}
where $\left \langle \cdot , \cdot \right \rangle $ denotes the inner product. $m,n \in [1,V]$. To keep the consistency for the similar but different samples in local cross-view, we align the pseudo label graph $\textbf{W} \in \mathbb{R}^{N \times N}$ and view feature graph $\textbf{S} \in \mathbb{R}^{N \times N}$ as:
\begin{equation} 
\begin{aligned} 
\mathcal{L}_{local} = \sum_{b=1}^B ( {-\textbf{W}_{ii}^blog( \frac{exp{(\textbf{S}_{ii}^b)}}{\sum exp({\textbf{S}_{ij}^b})}) } \\
- \sum_{j=1,j \neq i}^N \textbf{W}_{ij}^b log ( \frac{exp{(\textbf{S}_{ij}^b)}}{\sum exp({\textbf{S}_{ij}^b})})),
\end{aligned}
\label{local_con}
\end{equation}


Since the local cross-view features are constructed by the pair-wise view feature, there are $C^2_v$ combinations, thus the value of $B$ is set to $C^2_v$.

In Eq.\eqref{local_con}, the first term encourages the same sample in different views to generate similar features. It is the self-loops in constructed pseudo-label graph. Moreover, the second term encourages different samples with similar pseudo labels in different views to have similar features, thus keeping the consistency in the cross-view scenario.

\subsection{Objective Function}

For the multi-view data, it is common for the label information to be kept consistent. Following this assumption, we design a Mean Squared Error loss to keep the consistent pseudo-label graph on local and global levels:

\begin{equation}
\mathcal{L}_{con} = \frac{1}{B}{\sum_{b=1}^{B} {||\textbf{W}^G- \textbf{W}^{b}||^2_F}},
\label{con_Loss}
\end{equation}
where $\textbf{W}^G$ and $\textbf{W}^{b}$ is the global and local pseudo-label graph, respectively. Besides, it has $C^2_v$ combinations for pair-wise features. Thus the value of B is $C^2_v$.

The objective function of proposed DealMVC contains the reconstruction loss $\mathcal{L}_R$, the local contrastive calibration loss $\mathcal{L}_{local}$, the global contrastive calibration loss $\mathcal{L}_{global}$, and the pseudo-label consistency loss $\mathcal{L}_{con}$. In summary, the objective of DealMVC is formulated as follows:
\begin{equation}
\mathcal{L} = \mathcal{L}_{R}+ \alpha \mathcal{L}_{local} + \beta \mathcal{L}_{global} + \mu \mathcal{L}_{con} ,
\label{total_Loss}
\end{equation}
where $\alpha$, $\beta$ and $\mu$ are the trade-off hyper-parameters. The detailed learning process of our proposed DealMVC is shown in Algorithm \ref{ALGORITHM}.

\begin{table}[t]
\centering
\caption{Statistics summary of eight datasets.}
\scalebox{1.}{
\begin{tabular}{@{}clclclc@{}}
\toprule
Dataset      &  & Samples &  & Clusters &  & Views \\ \midrule
BBCSport     &  & 544     &  & 5        &  & 2     \\
Reuters      &  & 1200    &  & 6        &  & 5     \\
Caltech101\_7 &  & 1400    &  & 7        &  & 5     \\
Cora         &  & 2708    &  & 7        &  & 4     \\
Wiki         &  & 2866    &  & 10       &  & 2     \\
Caltech101   &  & 9144    &  & 102      &  & 5     \\
Hdigit       &  & 10000   &  & 10       &  & 2     \\
STL10        &  & 13000   &  & 10       &  & 4     \\ \bottomrule
\end{tabular}}
\label{DATASET_INFO} 
\end{table}

\begin{table*}[]
\centering
\caption{Clustering performance across eight multi-view benchmark datasets (Part 1/2). The most outstanding results are denoted in \textbf{bold}, while the second-best values are \underline{underlined}.}
\scalebox{1.15}{
\begin{tabular}{c|ccc|ccc|ccc|ccc}
\hline
\textbf{Methods}            & \multicolumn{3}{c|}{\textbf{BBCSport}}              & \multicolumn{3}{c|}{\textbf{Reuters}}               & \multicolumn{3}{c|}{\textbf{Caltech101\_7}}         & \multicolumn{3}{c}{\textbf{Cora}}                   \\ \hline
\textbf{Metrics (\%)} & \textbf{ACC}    & \textbf{NMI}    & \textbf{PUR}    & \textbf{ACC}    & \textbf{NMI}    & \textbf{PUR}    & \textbf{ACC}    & \textbf{NMI}    & \textbf{PUR}    & \textbf{ACC}    & \textbf{NMI}    & \textbf{PUR}    \\ \hline
\textbf{AE2-Net\cite{AE2Net}}           & 35.48          & 20.00          & 32.99          & 22.55          & 3.480          & 24.42          & 49.72          & 36.09          & 52.05          & 22.15          & 10.79          & 30.23          \\
\textbf{COMIC\cite{COMIC}}              & 32.28          & 21.51          & 35.69          & 16.67          & 12.57          & 16.67          & 53.20          & 54.90          & 60.40          & 30.21          & 21.47          & 30.21          \\
\textbf{DEMVC\cite{DEMVC}}              & 71.88          & {\ul 58.45}    & 75.47          & {\ul 45.00}    & {\ul 21.08}    & {\ul 45.08}    & 54.86          & 41.77          & 56.00          & 30.54          & 16.34          & 34.56          \\
\textbf{SDMVC\cite{SDMVC}}              & 51.29          & 32.86          & 57.23          & 17.67          & 14.73          & 18.58          & 44.21          & 30.97          & 47.93          & 31.06          & 15.12          & 32.05          \\
\textbf{SDSNE\cite{SDSNE}}              & {\ul 76.62}    & 57.50          & {\ul 75.84}    & 23.25          & 20.28          & 27.00          & {76.14}          & {\ul 74.52}    & 78.93          & {\ul 44.24}    & {\ul 34.35}    & {\ul 46.90}    \\
\textbf{DSMVC\cite{DSMVC}}              & 41.91          & 14.67          & 47.59          & 43.83          & 18.11          & 45.00          & {62.64}    & 47.34          & {64.43}    & 28.88          & 18.14          & 36.04          \\
\textbf{CoMVC\cite{SiMVC}}              & 37.50          & 19.82          & 40.81          & 32.25          & 13.58          & 33.83          & 40.84          & 28.66          & 69.95          & 29.76          & 4.640          & 33.94          \\
\textbf{SiMVC\cite{SiMVC}}              & 37.87          & 10.99          & 40.07          & 33.58          & 10.36          & 33.67          & 75.03          & 37.07          & 77.54          & 23.08          & 3.010          & 30.61          \\
\textbf{MFLVC\cite{MFLVC}}              & 40.99          & 27.79          & 61.46          & 39.92          & 20.01          & 41.42          & {\ul 80.40}          & 70.30          & {\ul 80.40}          & 31.02          & 12.97          & 39.22          \\ \hline
\textbf{DealMVC}                & \textbf{80.70} & \textbf{65.59} & \textbf{80.70} & \textbf{47.05} & \textbf{26.32} & \textbf{48.36} & \textbf{88.71} & \textbf{80.95} & \textbf{88.71} & \textbf{49.07} & \textbf{37.75} & \textbf{60.67} \\ \hline
\end{tabular}}
\label{com_tab1}
\end{table*}

\begin{table*}[]
\centering
\caption{Clustering performance across eight multi-view benchmark datasets (Part 2/2). The most exceptional results are marked in \textbf{bold}, and the second-best values are \underline{underlined}. The notation OOM signifies an out-of-memory error encountered during the training process.}
\scalebox{1.15}{
\begin{tabular}{c|ccc|ccc|ccc|ccc}
\hline
\textbf{Methods}            & \multicolumn{3}{c|}{\textbf{Wiki}}                  & \multicolumn{3}{c|}{\textbf{Hdigit}}                & \multicolumn{3}{c|}{\textbf{Caltech101}}            & \multicolumn{3}{c}{\textbf{STL10}}                  \\ \hline
\textbf{Metrics (\%)} & \textbf{ACC}    & \textbf{NMI}    & \textbf{PUR}    & \textbf{ACC}    & \textbf{NMI}    & \textbf{PUR}    & \textbf{ACC}    & \textbf{NMI}    & \textbf{PUR}    & \textbf{ACC}    & \textbf{NMI}    & \textbf{PUR}    \\ \hline
\textbf{AE2-Nets\cite{AE2Net}}           & 45.33          & 43.04          & 52.41          & 83.74          & 86.56          & 90.81          & 13.95          & {\ul 30.54}          & 25.43          & 9.600          & 17.76          & 23.19          \\
\textbf{COMIC\cite{COMIC}}              & 43.77          & {\ul 53.52}          & 58.57          & 94.21          & 86.96          & 94.21          & 17.22          & 26.88          & 20.67          & 18.27          & 11.76          & 18.71          \\
\textbf{DEMVC\cite{DEMVC}}              & 25.44          & 24.09          & 31.26          & 38.65          & 37.08          & 42.17          & 11.05          & 22.84          & 19.66          & 28.34          & {\ul 26.19}    & 30.13          \\
\textbf{SDMVC\cite{SDMVC}}              & 24.08          & 11.78          & 27.11          & 27.86          & 23.09          & 28.13          & 15.11          & 30.48 & {\ul 28.78}    & 30.01          & 25.27          & 31.07          \\
\textbf{SDSNE\cite{SDSNE}}              & {56.07}    & 52.68    & {\ul 61.83}    & O/M             & O/M             & O/M             & O/M             & O/M             & O/M             & O/M             & O/M             & O/M             \\
\textbf{DSMVC\cite{DSMVC}}              & {\ul 58.53}          & \textbf{54.60} & 60.15          & 98.06    & {97.36}    & 98.06    & { 16.27}    & 26.53          & 25.60          & {27.53}    & 19.33          & {29.41}    \\
\textbf{CoMVC\cite{SiMVC}}              & 26.94          & 26.24          & 29.03          & 34.63          & 39.81          & 36.35          & 16.36          & 25.61          & 23.71          & 23.55          & 16.26          & 25.01          \\
\textbf{SiMVC\cite{SiMVC}}              & 21.74          & 7.030          & 22.16          & 23.48          & 11.20          & 25.02          & 13.48          & 18.18          & 17.91          & 16.04          & 6.340         & 17.02          \\
\textbf{MFLVC\cite{MFLVC}}              & 48.08          & {41.88} & 50.87          & {\ul 98.82}          & {\ul 98.41}         & {\ul 98.82}          & {\ul 21.30}          & 28.60          & 28.23          & {\ul 31.14}          & 25.36          & {\ul 31.25}          \\ \hline
\textbf{DealMVC}                & \textbf{59.28} & 52.36          & \textbf{62.21} & \textbf{99.80} & \textbf{99.34} & \textbf{99.80} & \textbf{22.26} & \textbf{30.77}    & \textbf{29.53} & \textbf{36.44} & \textbf{29.93} & \textbf{36.95} \\ \hline
\end{tabular}}
\label{com_tab2}
\end{table*}

\section{EXPERIMENT}
\subsection{Dataset}
To ascertain the efficacy of our proposed DealMVC method, we carry out comprehensive experiments across eight benchmark datasets: BBCSport, Reuters, Caltech101\_7, Cora, Wiki, Hdigit, Caltech101, and STL10. A succinct overview of these datasets is presented in Table \ref{DATASET_INFO}. In this section, we implement experiments to verify the effectiveness of the proposed DealMVC through answering the following questions:

\begin{itemize}
\item \textbf{RQ1}: How effective is DealMVC for deep multi-view clustering?
\item \textbf{RQ2}: How does the proposed module influence the performance of DealMVC?
\item \textbf{RQ3}: How do the hyper-parameters impact the performance of DealMVC?
\item \textbf{RQ4}: What is the clustering structure revealed by DealMVC?
\end{itemize}

\begin{table*}[]
\centering
\caption{Ablation studies of the proposed strategies. ``(w/o) $\textbf{L}$'', ``(w/o) $\textbf{G}$'' and ``(w/o) $\textbf{L}\&\textbf{G}$ '' represent the reduced models by removing the local contrastive calibration, the global contrastive calibration and both, respectively. Moreover,  ``(w/o) $\textbf{S}$'' is the model by removing the pseudo-label consistency loss. Similarly, ``(w/o) $\textbf{P}$'', ``(w/o) $\textbf{A}$'' and ``(w/o) $\textbf{P}\&\textbf{A}$ '' denote the reduced models by removing the learnable sampling network, the attention mechanism, and both, respectively.}
\scalebox{1.1}{
\begin{tabular}{c|c|cccc|ccc|c}
\hline
\textbf{Datasets}                      & \textbf{Metrics (\%)} & \textbf{(w/o)L} & \textbf{(w/o)G} & \textbf{(w/o)L\&G} & \textbf{(w/o)S} & \textbf{(w/o)P} & \textbf{(w/o)A} & \textbf{(w/o)P\&A} & \textbf{Ours} \\ \hline
\multirow{3}{*}{\textbf{BBCSport}}     & ACC              & 64.96          & 52.76          & 41.18             & 54.06          & 53.13          & 53.47          & 51.21             & \textbf{80.70}        \\
                                       & NMI              & 46.79          & 36.18          & 15.92             & 34.79          & 34.88          & 35.17          & 32.58             & \textbf{65.59}        \\
                                       & PUR              & 68.01          & 58.01          & 45.28             & 59.54          & 59.12          & 58.57          & 57.00             & \textbf{80.70}        \\ \hline
\multirow{3}{*}{\textbf{Reuters}}      & ACC              & 41.82          & 50.21          & 40.97             & 49.63          & 36.92          & 45.42          & 41.25             & \textbf{47.05}        \\
                                       & NMI              & 20.59          & 31.57          & 17.72             & 30.61          & 17.08          & 27.51          & 24.61             & \textbf{26.32}        \\
                                       & PUR              & 42.42          & 51.48          & 41.85             & 59.92          & 37.67          & 47.92          & 44.08             & \textbf{48.36}        \\ \hline
\multirow{3}{*}{\textbf{Caltech101\_7}} & ACC              & 41.09          & 86.64          & 80.50             & 84.64          & 85.07          & 85.79          & 84.86             & \textbf{88.71}        \\
                                       & NMI              & 41.60          & 78.22          & 72.87             & 75.68          & 77.68          & 77.29          & 75.87             & \textbf{80.95}        \\
                                       & PUR              & 41.71          & 86.64          & 80.50             & 84.64          & 85.07          & 85.79          & 84.86             & \textbf{88.71}        \\ \hline
\multirow{3}{*}{\textbf{Cora}}         & ACC              & 33.25          & 41.07          & 28.92             & 40.01          & 39.90          & 37.57          & 42.96             & \textbf{49.07}        \\
                                       & NMI              & 15.66          & 25.10          & 7.990             & 24.45          & 22.62          & 19.82          & 25.53             & \textbf{37.75}        \\
                                       & PUR              & 37.25          & 46.10          & 34.47             & 45.75          & 44.89          & 43.44          & 47.43             & \textbf{60.67}        \\ \hline
\multirow{3}{*}{\textbf{Wiki}}         & ACC              & 40.13          & 53.88          & 38.27             & 53.10          & 53.97          & 52.82          & 54.38             & \textbf{59.28}        \\
                                       & NMI              & 31.42          & 47.01          & 25.52             & 47.43          & 46.74          & 45.98          & 51.31             & \textbf{52.36}        \\
                                       & PUR              & 41.21          & 56.48          & 43.02             & 57.08          & 56.61          & 55.49          & 59.27             & \textbf{62.21}        \\ \hline
\multirow{3}{*}{\textbf{Hdigit}}       & ACC              & 60.98          & 90.67          & 41.13             & 88.66          & 91.74          & 94.69          & 89.79             & \textbf{99.80}        \\
                                       & NMI              & 73.36          & 95.84          & 38.53             & 94.61          & 96.32          & 97.21          & 95.55             & \textbf{99.34}        \\
                                       & PUR              & 61.27          & 90.70          & 43.28             & 88.69          & 91.74          & 94.72          & 89.80              & \textbf{99.80}        \\ \hline
\multirow{3}{*}{\textbf{CALTECH101}}   & ACC              & 21.88          & 22.26          & 18.08             & 20.67          & 21.22          & 20.61          & 21.23             & \textbf{22.26}        \\
                                       & NMI              & 21.88          & 22.60          & 24.28             & 33.39          & 28.97          & 29.85          & 24.25             & \textbf{30.77}        \\
                                       & PUR              & 28.92          & 28.52          & 24.13             & 27.72          & 27.43          & 27.93          & 27.76             & \textbf{29.53}        \\ \hline
\multirow{3}{*}{\textbf{STL10}}        & ACC              & 18.78          & 21.52          & 16.24             & 29.97          & 32.34          & 24.73          & 28.43             & \textbf{36.44}        \\
                                       & NMI              & 17.10          & 14.72          & 9.720             & 25.80          & 27.94          & 19.60          & 23.32             & \textbf{29.93}        \\
                                       & PUR              & 17.54          & 22.00          & 16.29             & 30.71          & 33.32          & 25.06          & 29.23             & \textbf{36.95}        \\ \hline
\end{tabular}}
\label{ablation_study}
\end{table*}

\begin{figure*}[!ht]
\vspace{20pt}
\begin{center}
{
\centering
\subfloat[Raw Features]{{\includegraphics[width=0.22\textwidth]{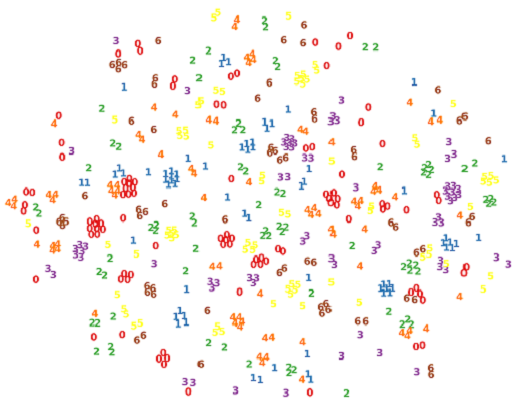}} \label{figure_3a}}\hspace{2mm}
\subfloat[Final Result]{{\includegraphics[width=0.22\textwidth]{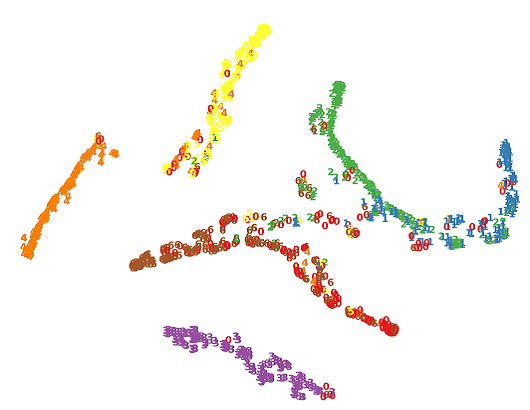}} \label{figure_3c}}\hspace{2mm}
\subfloat[Raw Features]{{\includegraphics[width=0.22\textwidth]{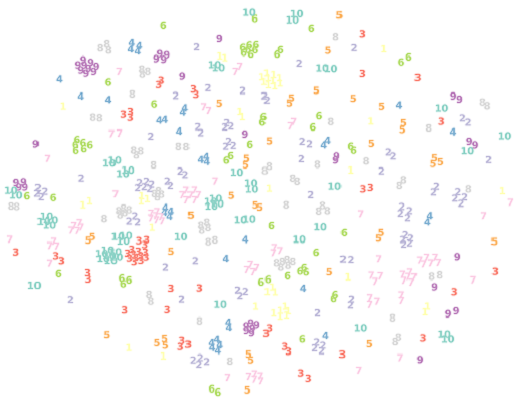}} \label{figure_3d}}\hspace{2mm}
\subfloat[Final Result]{{\includegraphics[width=0.22\textwidth]{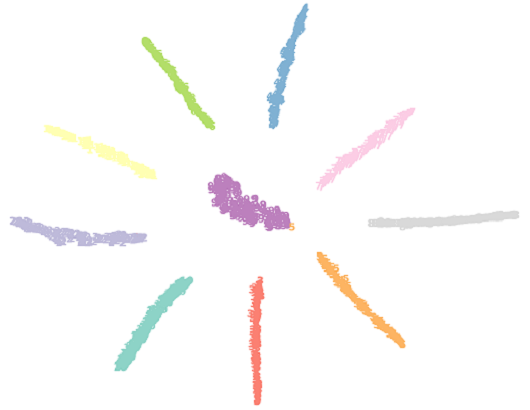}} \label{figure_4d}}
\vspace{3 pt}
\caption{Visualization the feature on Caltech101\_7 and Hdigit datasets.}
\label{t_sne}
}
\end{center}
\end{figure*}

\subsection{Experiment Setup}
The experiments are executed using the following hardware configuration: Intel Core i7-7820x CPU, NVIDIA GeForce RTX 3080 GPU, and 64GB RAM. Additionally, the PyTorch platform is employed for all experiments. In the case of DealMVC, we utilize the Adam \cite{ADAM} optimizer to minimize the total loss \eqref{total_Loss}.

\subsubsection{Comparison methods}
The proposed DealMVC is benchmarked against seven prominent deep multi-view clustering algorithms. Specifically, these compared clustering algorithms can be broadly categorized into two groups: deep multi-view clustering algorithms (AE2-Net \cite{AE2Net}, CoMIC \cite{COMIC}, DEMVC \cite{DEMVC}, SDMVC \cite{SDMVC}, SDSNE \cite{SDSNE}, DSMVC \cite{DSMVC}, SiMVC \cite{SiMVC}) and contrastive multi-view clustering algorithms (CoMVC \cite{SiMVC}, MFLVC \cite{MFLVC}).

\subsubsection{Parameter Settings}
For all baseline methods, we replicate results using their provided source code with the original configurations. In the case of our proposed DealMVC, a batch size of 256 is employed consistently across all datasets. The learning rate is set at 0.0003. We utilize the autoencoder as the pre-training model, assigning 300 epochs for pre-training and 100 epochs for contrastive learning. The trade-off hyper-parameters $\alpha$, $\beta$, and $\mu$ are uniformly set to 1.0.

\begin{figure*}[!htbp]
    \centering
    \includegraphics[width=1.0\textwidth]{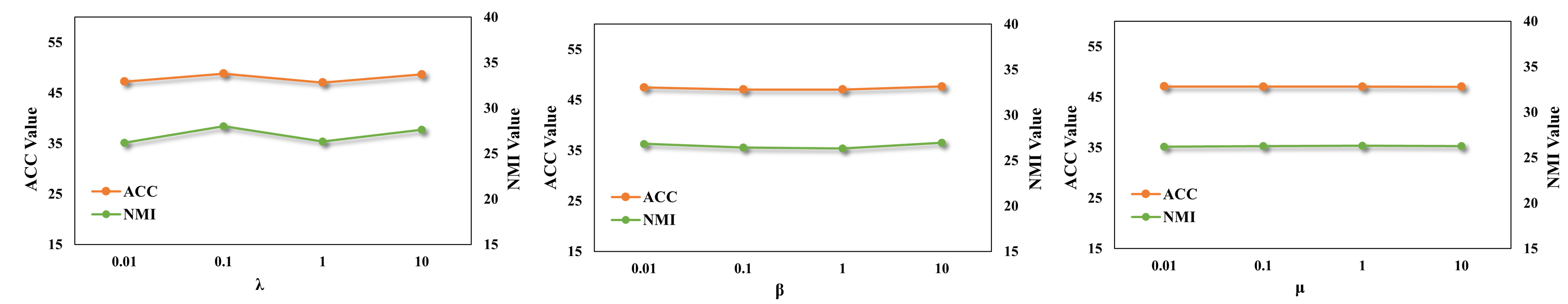}
    \caption{Sensitivity analysis of the hyper-parameters on Reuters dataset.}
    \label{sen}
\end{figure*}

\subsubsection{Evaluation metrics}
To evaluate the effectiveness and superiority of our DealMVC, we adopt the widely used metrics, i.e., clustering accuracy (ACC), normalized mutual information (NMI), and purity (PUR).\cite{ZHOU_1,siwei_1,siwei_2}.

\subsection{Performance Comparison~\textbf{(RQ1)}}

To demonstrate the superiority of our proposed DealMVC, we compare DealMVC with nine baselines, including deep multi-view clustering methods (AE2-Nets \cite{AE2Net}, COMIC \cite{COMIC}, DEMVC \cite{DEMVC}, SDMVC \cite{SDMVC}, SDSNE \cite{SDSNE}, DSMVC \cite{DSMVC}, CoMVC \cite{SiMVC}) and contrastive multi-view clustering methods (SiMVC \cite{SiMVC}, MFLVC \cite{MFLVC}). In Table \ref{com_tab1} and \ref{com_tab2}, the clustering performance of all the compared methods on the eight datasets is presented. From these results, we have the following observations. 1) In comparison to classical deep multi-view clustering algorithms, our DealMVC consistently achieves the most favorable clustering outcomes across the majority of datasets. To illustrate, when examining the outcomes on STL10, DealMVC surpasses its closest competitors by notable increments of 5.30\%, 3.74\%, and 5.70\% in terms of ACC, NMI, and PUR, respectively. We conjecture the reason is that those methods do not consider the contrastive learning mechanism. 2) Thanks to the local and global contrastive strategies, our method obtains better performance with contrastive deep multi-view clustering algorithms. 3) Our DealMVC achieve promising performance on a large dataset, i.e., Hdigit and STL10. The achieved results effectively showcase the generalization prowess of DealMVC. In summary, the aforementioned observations substantiate the exceptional performance of our proposed DealMVC.

\subsection{Ablation Studies~\textbf{(RQ2)}}

\subsubsection{Effectiveness of the adaptive global fusion module}\label{fusion}

To verify the effectiveness of the adaptive global fusion mechanism, we conduct experiments on eight datasets. For simplicity, we represent ``(w/o) $\textbf{P}$'', ``(w/o) $\textbf{A}$'' and ``(w/o) $\textbf{P}\&\textbf{A}$ '' as removing the learnable sampling network, the attention mechanism, and both, respectively. Here, we concatenate the view representation $\textbf{Z}^v$ directly as global view representation, i.e., ``(w/o) $\textbf{P}\&\textbf{A}$ ''. The results in Table \ref{ablation_study} reveal a notable trend: the clustering performance of DealMVC experiences a decline when any of the aforementioned components is omitted. The attention mechanism and the learnable sampling network are interacted and jointly optimized by two manners. Thus the global cross-view feature is more representative.

\subsubsection{Effectiveness of the dual contrastive calibration module} \label{local_global_sec}
In this section, we carry out experiments to validate the efficacy of the proposed modules. The corresponding results are presented in Table \ref{ablation_study}. Concretely, we adopt ``(w/o) $\textbf{L}$'', ``(w/o) $\textbf{G}$'' and ``(w/o) $\textbf{L}\&\textbf{G}$ '' to denote the reduced models by removing the local module, the global module, the both respectively. Here, the autoencoder model is regarded as the baseline, i.e., ``(w/o) $\textbf{L}\&\textbf{G}$ ''. From the results in Table. \ref{ablation_study}, we could observe that. 1) Without any of local or global contrastive calibration loss, the model performance will decrease, indicating that each module makes contributions to boosting the performance. 2) Since the training of feature structure is guided by reliable class information, the discriminative capacity of the learned features is guaranteed, thus achieving better performance. Overall, the aforementioned observations have verified the effectiveness of the proposed modules in our proposed DealMVC.

\subsubsection{Effectiveness of the proposed pseudo-label consistent loss}
We implement the experiments to validate the effectiveness of the proposed pseudo-label consistency loss. The results are shown in Table. \ref{ablation_study}. We can observe that when removing the pseudo-label loss, the multi-view clustering performance is limited. We analyze the reason as follows. The class information for the multi-view data should be consistent under local and global cross-view feature circumstances. The pseudo-label loss could keep the class information consistent by minimizing the mean squared error for the pseudo-label graph at both local and global levels.

\subsection{Hyper-parameter Analysis~\textbf{(RQ3)}}
To assess the robustness of our proposed method DealMVC concerning the trade-off hyperparameters, we conduct experiments using the Reuters dataset. As depicted in Fig. \ref{sen}, we discern a notable influence of $\alpha$ on the model's performance. Conversely, the impact of $\mu$ is relatively minor.

\subsection{Visualization Analysis~\textbf{(RQ4)}}

In this subsection, we conduct the visualization experiment to demonstrate the superiority of DealMVC intuitively. To be specific, we visualize the distribution of the learned embeddings of DealMVC on Caltech101\_7 and Hdigit datasets via $t$-SNE algorithm \cite{T_SNE}. From the results in Fig.\ref{t_sne}, we conclude that compared with the raw features, DealMVC could better reveals the intrinsic clustering structure.

\section{CONCLUSION}
In this paper, we design a novel dual contrastive calibration network for multi-view clustering, termed DealMVC. Specifically, we propose a fusion mechanism to obtain a global cross-view feature. Then, by aligning the view feature similarity graph and the high-confidence pseudo-label graph by the global contrastive calibration loss, similar but different samples could keep consistent in the cross-view feature. To further use the diversity of multi-view data, we design a local contrastive calibration loss to constrain the consistency of pair-wise view features. The structure of the cross-view local feature is regularized by the reliable class information, thus promising similar but different samples to have same cluster. During the training process, the cross-view feature is optimized at both local and global levels. Thorough experiments conducted on eight datasets convincingly demonstrate the effectiveness of our proposed multi-view clustering algorithm.

\begin{acks}
This work was supported by the National Key R$\&$D Program of China 2020AAA0107100 and the National Natural Science Foundation of China (project no. 62325604, 62276271). 
\end{acks}

\bibliographystyle{ACM-Reference-Format}
\newpage
\balance
\bibliography{ref}
\end{document}